\documentclass[11pt,a4paper]{article}

\usepackage[utf8]{inputenc} 
\usepackage{url}            
\usepackage{booktabs}       
\usepackage{amsfonts}       
\usepackage{nicefrac}       
\usepackage{amsmath}
\usepackage{float}
\usepackage{subcaption}

\usepackage[nocustomfonts]{nxai}
\definecolor{nxaiLinkBlue}{HTML}{3073AD}
\hypersetup{linkcolor=nxaiLinkBlue,citecolor=nxaiLinkBlue,urlcolor=nxaiLinkBlue}
\newcommand{\affmark}[1]{\raisebox{0.55ex}{\fontsize{7}{7}\selectfont #1}}

\providecommand{\doi}[1]{}
\renewcommand{\doi}[1]{doi: \href{https://doi.org/#1}{\nolinkurl{#1}}}

\ifdefined\appxMacrosLoaded\else
\newcommand{\appxMacrosLoaded}{}

\newcommand{\appxMlstmStateEntries}{32{,}768}
\newcommand{\appxSlstmStateEntries}{512}
\newcommand{\appxStateSizeRatio}{64}
\newcommand{\appxSpectralHeadParams}{2{,}196{,}480}
\newcommand{\appxSpectralHeadSlstmParams}{33{,}792}
\newcommand{\appxSpectralHeadMlstmParams}{2{,}162{,}688}
\newcommand{\appxTextHeadParams}{1{,}131{,}520}

\newcommand{\appxLoraPerRank}{96{,}768}

\newcommand{\appxFftCentreBins}{$1$, $2$, $3$, $4$, $5$, $6$, $7$, $8$, $9$, $10$, $11$, $13$, $14$, $16$, $18$, $20$, $22$, $25$, $28$, $31$, $35$, $39$, $43$, $49$, $54$, $61$, $68$, $76$, $84$, $94$, $105$, $118$, $132$, $147$, $164$, $184$, $205$, $229$, $256$, $286$, $320$, $357$, $399$, $446$, $498$, $556$, $622$, $695$, $776$, $867$, $969$, $1{,}083$, $1{,}210$, $1{,}351$, $1{,}510$, $1{,}687$, $1{,}885$, $2{,}106$, $2{,}353$, $2{,}629$, $2{,}937$, $3{,}282$, $3{,}667$ and $4{,}096$}
\newcommand{\appxTextEmbedDim}{1{,}024}
\newcommand{\appxSeriesEmbedDim}{6{,}144}
\newcommand{\appxSharedDim}{32}

\newcommand{\appxClipTextWidths}{1{,}024 \to 256 \to 256 \to 256 \to 32}
\newcommand{\appxClipSeriesWidths}{6{,}144 \to 32}
\newcommand{\appxClipTextParams}{402{,}208}
\newcommand{\appxClipSeriesParams}{196{,}640}
\newcommand{\appxClipTotalParams}{598{,}848}

\newcommand{\appxGiftevalSteps}{50{,}000}
\newcommand{\appxGiftevalSamplesM}{12.8}

\fi

\newenvironment{ack}{\section*{Acknowledgments and Disclosure of Funding}}{}

\nxaisetalogo{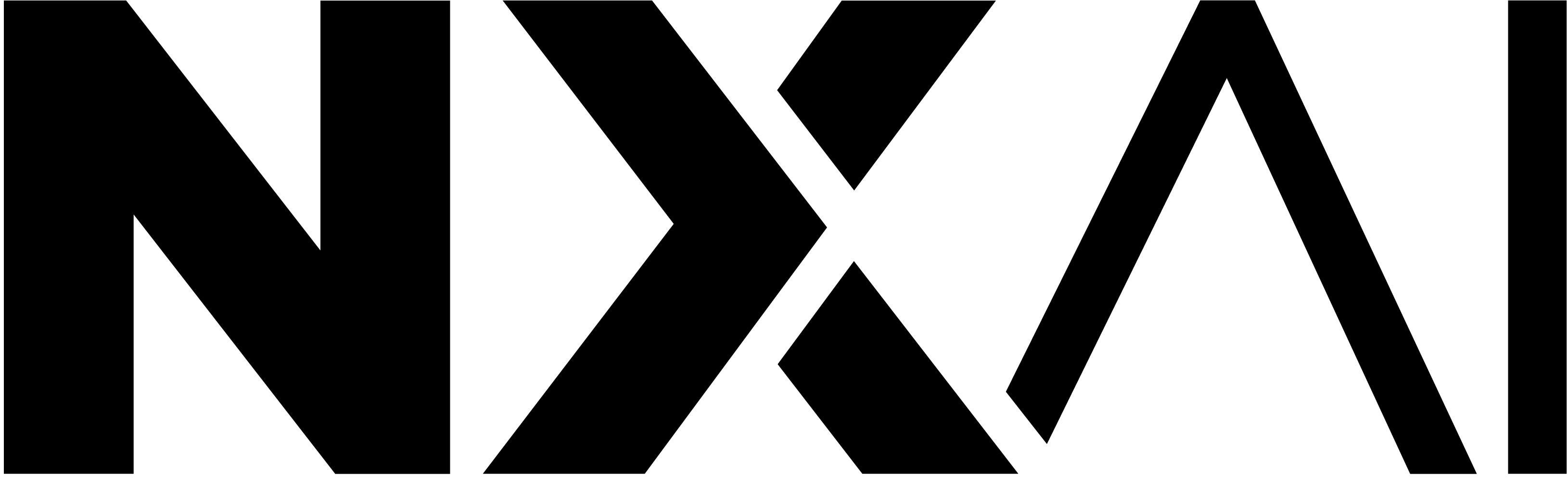}
\nxaisetlogoheight{4mm}
\nxaisetdate{}

\title{The Unused Input: The Initial State as a Conditioning Interface for Recurrent TSFMs}
\author{Bernhard Voggenberger\affmark{1,2,*}\and Christian Ganh\"{o}r\affmark{1}\and Levente Z\'{o}lyomi\affmark{1,2}\and Sebastian B\"{o}ck\affmark{1}\and Sepp Hochreiter\affmark{1,2}}
\nxaisetaffiliations{%
\affmark{1}NXAI GmbH, Linz, Austria \quad \affmark{2}Institute for Machine Learning, Johannes Kepler University, Linz, Austria
}
\nxaisetfooter{Accepted at the NeurIPS 2026 Workshop on Foundation Models for Temporal Systems (FMTS).\\
Corresponding author: Sebastian B\"{o}ck (\texttt{sebastian.boeck@nx-ai.com}) \quad \affmark{*}Work done during an internship at NXAI GmbH.}

\nxaiabstract{%
Time series foundation models (TSFMs) forecast unseen series without task-specific training. Increasingly they are built on recurrent backbones, whose constant per-step cost makes long-context inference cheap, and which, unlike transformers, carry a state.
    A recurrent backbone computes its prediction from two inputs: the observed context window and the initial recurrent state. In practice only the first is utilised and the initial state is unused by convention. We ask whether the initial state can be utilised in recurrent TSFMs. Using the xLSTM-based zero-shot forecaster TiRex as a case study, we freeze the backbone and train only a small head that writes a context vector into the initial state, so that any change in accuracy is attributable to the injected state rather than to adapted weights. A spectral summary of the input window, learned once and applied unchanged to every series, lowers mean GiftEval MASE by $0.175\%$, with every seed improving, which neither a parameter-matched LoRA adapter nor a learnable static state reproduces. An embedding of text attached to each series, evaluated on six MoTime datasets, reveals two roles for the state. As a per-dataset adaptation surface, a static state lowers MASE by $0.04$--$6.96\%$ of the zero-shot baseline ($8$--$70\%$ of the gain of full fine-tuning). As a channel for exogenous information, the text-conditioned state lowers it by a further $0.15$--$0.41\%$ of the zero-shot baseline on four of six datasets and, combined with a LoRA adapter, exceeds full fine-tuning, which has no access to the text, on one of them. In TiRex, the effect concentrates in the mLSTM blocks and is strongest at short context. The results are preliminary and the improvements small, but they point to the initial state as a direction for further research.
}

\makeatletter
\ifcsname iflatexml\endcsname\else\expandafter\newif\csname iflatexml\endcsname\fi
\newcommand{\nxaihtmlfrontmatter}{%
  \iflatexml
    \par\noindent\nxai@affiliations\par
    \par\noindent\nxai@footertext\par
    \section*{Abstract}\nxai@abstract\par
  \fi}
\makeatother

\begin{document}
\maketitle
\nxaihtmlfrontmatter

\section{Introduction}
Time series foundation models, pretrained once and applied across many forecasting tasks without further training, have become a standard tool for zero-shot forecasting \citep{pmlr-v235-das24c, ansari2024chronos, pmlr-v235-woo24a, cohen2024tototimeseriesoptimized}. Many are built on recurrent architectures, like TiRex \citep{auer2025tirexzeroshotforecastinglong, podest2026tirex2generalizingtirexmultivariate} on xLSTM \citep{beck2024xlstmextendedlongshortterm}, TSMamba \citep{ma2024mambafoundationmodeltime} and Mamba4Cast \citep{bhethanabhotla2024mambacast} on Mamba, and others on gated DeltaNet \citep{hartl2026subquadraticarchitecturesapplicationsprinciples}. A recurrent forecaster produces its forecast as a function of two inputs: the context window it observes, and the initial state it starts from. In practice only the first is used; the initial state is set to a fixed default that holds no information about the series being forecast. We investigate this unused initial state, taking TiRex as a case study. 
 
Prompt and prefix tuning adapt a frozen model by optimising a sequence of trained vectors that its inputs attend to, leaving its weights unchanged \citep{lester-etal-2021-power, li-liang-2021-prefix}; recent work shows that similar conditioning mechanisms extend to time series forecasters \citep{ahamed2026reasoningawaretrainingtimeseries, wang2025chronosteerbridginglargelanguage}. Initialising the state of classical RNNs is known to be beneficial, both for forecasting and systems-modelling tasks \citep{7966138, 8630673}, and for conditioning on context more broadly \citep{wenke2019contextualrecurrentneuralnetworks}, and the Temporal Fusion Transformer initialises its LSTM from static covariates \citep{lim2021temporal}. In NLP, \citet{kang-etal-2025-state} extend this idea to modern state space models, adapting them by directly tuning states rather than prepending prompts, and \citet{pmlr-v267-galim25a} show that, for state space models, tuning the initial state of a frozen model is at least as expressive as prefix tuning. To our knowledge, we are the first to study the initial state of a frozen, pretrained recurrent time series foundation model as a conditioning interface, and to characterise how the injected state behaves.
 
Our contributions are: (1) using TiRex, we show the recurrent initial state is a usable conditioning interface that can improve a frozen time series foundation model; (2) we condition on a spectral summary of the input window and on a CLIP-aligned \citep{pmlr-v139-radford21a} embedding of exogenous text, comparing each against a parameter-matched LoRA adapter \citep{hu2022lora} and an expressivity-matched static state; and (3) we characterise the behaviour of the injected state, whose influence in TiRex is strongest at short context and diminishes as data accumulates, and localises to the matrix memory of its underlying xLSTM (the mLSTM blocks).

\section{State conditioning}
\label{sec:setup}

\paragraph{Backbone and interface.}
The backbone follows the TiRex architecture and pretraining data mixture \citep{auer2025tirexzeroshotforecastinglong}, and is built on xLSTM. It interleaves two block types: sLSTM blocks with a scalar state and mLSTM blocks with a matrix state. The configuration deviates from the released TiRex checkpoint in the block arrangement, so that the effect of the initial state can be measured on both a scalar and a matrix state. We maintain a 1:1 ratio of the two blocks as it has shown to be a balanced tradeoff for this architecture family \citep{zolyomi2026nonlinear}. We train with a context window of $8{,}192$ steps, so that at patch size $32$ the model performs $256$ recurrent steps, enough to observe how the injected state decays over a long recurrence. At inference the model is normally run from an all-zero initial state $c_0=0$. We keep the backbone frozen and train only a small \emph{state-initialisation head} that writes a context vector into $c_0$ (one shared dense projection for all sLSTM blocks, one for all mLSTM). The head is small relative to the $38.4$M-parameter frozen backbone, its exact size differs by regime and is given below, and the controls are matched to its parameter budget (Appendix~\ref{app:arch}).

\paragraph{Spectral conditioning (window only).}
The first regime derives $c_0$ from the input window alone. We summarise the window's magnitude spectrum into a fixed $64$-dimensional vector, a compact descriptor of its dominant periodicities, and write it into $c_0$ through a small linear head ($2.2$M parameters, $5.7\%$ of the backbone). The head is trained once on the same data mix used to pretrain the checkpoint, never on the evaluation series, so one spectrum-to-$c_0$ map serves all series (Appendix~\ref{app:fft}).

\paragraph{Exogenous conditioning (text).}
The second regime draws on data outside the window. We use MoTime \citep{zhou2025motimedatasetsuitemultimodal}, a multimodal suite that pairs each series with an associated text, and retain six text-based evaluation sets drawn from five sources: MovieLens \citep{movielense}, News-FO \citep{wang2024newsforecast}, Tianchi \citep{tianchi2018rectmall}, Wiki People \citep{wikipeople}, and two splits of PixelRec (Full and Sub) \citep{cheng2023imagedatasetbenchmarkingrecommender}. Per dataset we align the series and its text in a shared space with a CLIP-style projection: the text is embedded with Qwen3-Embedding-0.6B \citep{qwen3embedding} and the series with a pretrained TiRex feature extractor \citep{auer2025pretrained}. The resulting text embedding is the exogenous context vector, written into $c_0$ by a shared linear head ($1.1$M parameters, $2.9\%$ of the backbone) as in the spectral regime; the alignment projection is trained separately ($0.6$M) and frozen, so it is not counted in the head's budget (Appendix~\ref{app:motime}).

\paragraph{Controls.}
The head adds parameters, so we separate the effect of \emph{content} from the effect of \emph{capacity} with two controls trained under identical conditions. A \emph{parameter-matched LoRA} adapter carries no context vector and adapts the backbone instead; each rank is set just above the head it is matched against (rank $23$, $2.23$M, against the $2.20$M spectral head; rank $12$, $1.16$M, against the $1.13$M text head). An \emph{expressivity-matched static state} ($0.2$M) replaces the content-dependent head with a per-layer learnable constant $c_0$ that receives no input, controlling for the value of a non-zero, learnable start. We also train head and LoRA jointly, against a parameter-equivalent LoRA-only run, to test whether the two effects stack.

\paragraph{Training.}
All heads and controls are trained with the backbone frozen (details in Appendix~\ref{app:eval}); a full fine-tuning run (all weights unfrozen) is included as a reference. Every trained configuration uses six seeds and no test-time augmentation, and we report MASE. On GiftEval the spectral head is evaluated zero-shot across the benchmark \citep{aksu2024giftevalbenchmarkgeneraltime}. On MoTime the text heads are trained per dataset on a training split and evaluated on a held-out split over fixed evaluation windows.

\section{Results}
\label{sec:results}
 
\begin{figure}[t]
\centering
\includegraphics[width=0.49\linewidth]{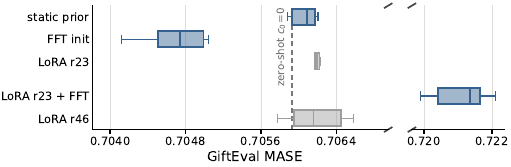}\hfill
\includegraphics[width=0.49\linewidth]{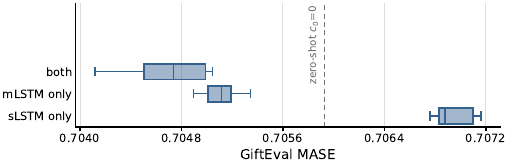}
\caption{GiftEval MASE, lower is better; blue uses a state initialisation, grey uses LoRA. Left: FFT init against parameter- and expressivity-matched controls. Right: FFT init into sLSTM, mLSTM, or both. Numbers in Appendices~\ref{app:gifteval} and~\ref{app:blocktype}.}
\label{fig:gifteval}
\end{figure}

\paragraph{GiftEval (spectral).}
Writing the window's own spectrum into $c_0$ lowers GiftEval MASE from a single-seed zero-shot baseline of $0.7059$ to a six-seed mean of $0.7047$, a $0.175\%$ relative improvement, with the backbone frozen and every seed improving (Figure~\ref{fig:gifteval}). The expressivity-matched static state ($0.7061$), the parameter-matched LoRA-r23 adapter ($0.7063$), a doubled-rank LoRA-r46 ($0.7062$) and LoRA trained jointly with the FFT head ($0.7211$) all remain above the baseline. None of the controls reproduces the improvement. Ablating by block type, injecting the spectrum into only the sLSTM blocks, only the mLSTM blocks, or both, places the effect in the mLSTM: mLSTM-only injection recovers most of the gain, while sLSTM-only injection is above the baseline.

\begin{figure*}[t]
\centering
\includegraphics[width=\linewidth,height=0.22\textheight,keepaspectratio]{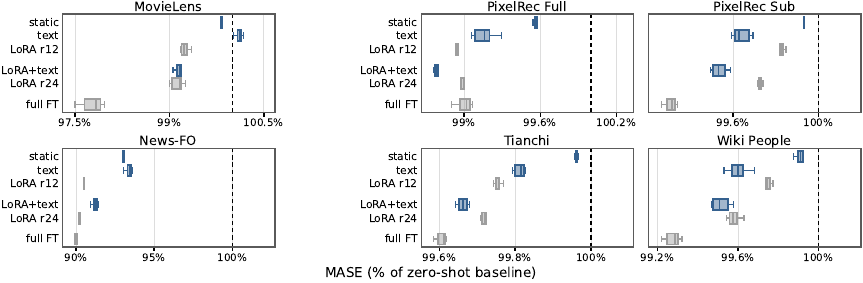}
\caption{MoTime MASE as a percentage of the zero-shot baseline, lower is better; blue uses a state initialisation, grey uses LoRA or full fine-tuning. Right: the four datasets where text beats static and LoRA+text beats LoRA r24. Left: the two tasks where text does not beat controls. Per-dataset numbers in Appendix~\ref{app:results}.}
\label{fig:motime_methods}
\end{figure*}

\paragraph{MoTime (exogenous text).}
We build the conditioning up from the weakest possible signal (Figure~\ref{fig:motime_methods}). The static state is a per-dataset constant: it encodes only which dataset a series belongs to, nothing about the series itself. Even this minimal signal improves the forecast on every dataset: the static state is below the zero-shot baseline on all six. On GiftEval, by contrast, a single static state derived from the pretraining corpus encodes no such distinction and does not improve on the baseline (Figure~\ref{fig:gifteval}). So the question is not whether the state helps, but how much the associated text adds beyond dataset identity. The text-conditioned state beats the static state on four of six datasets: both splits of PixelRec, Tianchi, and Wiki People, but not MovieLens or News-FO. Of those four, a matched LoRA adapter, carrying no text at all, is stronger on two (PixelRec Full and Tianchi), so at this budget capacity alone can substitute for the text. Combining the two recovers all four: at $2.2$M trainable parameters, the text-conditioned state with a LoRA adapter ($1.1$M each) beats a LoRA adapter of the same total size, and is the strongest parameter-efficient method on all four, though on Wiki People the margin is within seed variation. A shuffled-text control (Appendix~\ref{app:shuffled}) confirms the use of series-specific text on Tianchi, but not on Wiki People, whose aligned text embeddings are nearly identical across series.

Because the static state is a per-dataset adaptation, it is directly comparable to full fine-tuning, which adapts the same model to the same dataset. On that comparison the static state recovers between $8\%$ and $70\%$ of the fine-tuning gain, lowering MASE by $0.04\%$ on Tianchi and by $6.96\%$ on News-FO. The text-conditioned state is not a substitute for fine-tuning but an additional input, and it helps precisely where fine-tuning has least room: on the four datasets where it beats the static state, full fine-tuning itself improves zero-shot by only $0.39\%$ to $1.00\%$. Within that margin the combined text-conditioned state and LoRA model recovers $67\%$ to $121\%$ of the fine-tuning gain, exceeding it on PixelRec Full. 
 
\begin{figure}[t]
\centering
\begin{minipage}[b]{0.49\linewidth}
    \centering
    \includegraphics[width=\linewidth]{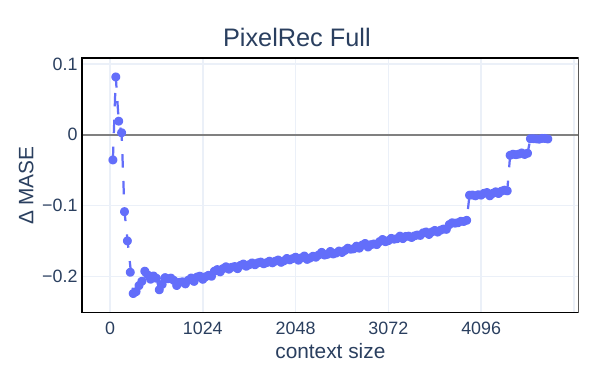}
\end{minipage}\hfill
\begin{minipage}[b]{0.49\linewidth}
    \centering
    \includegraphics[width=\linewidth]{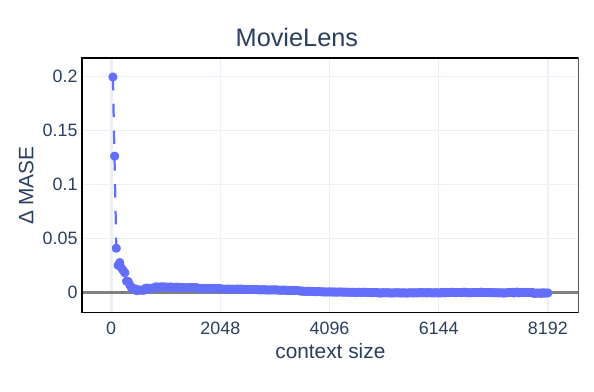}
\end{minipage}
\caption{Per dataset change in MASE over context length when injecting a text-conditioned initial state; negative values indicate improvement. On PixelRec Full the text-conditioned state beats the static state, on MovieLens it does not. The initial state's influence shrinks as context accumulates. Static and text-conditioned state decay curves for all datasets in Appendix~\ref{app:static-decay}.}
\label{fig:decay}
\end{figure}

We next measure the effect of the state alone. We give a fixed target progressively more context and forecast it with and without the text-conditioned initial state; the difference between the two is the effect of the state at that context length. Its influence is largest at short context and, after some instability at the very shortest windows, decays steadily as the window grows, both on PixelRec Full, where the text-conditioned state beats the static state, and on MovieLens, where it does not (Figure~\ref{fig:decay}).
We attribute this decay to two causes: the mLSTM's forget gate gradually overwrites the injected state, and a longer window leaves the already-stronger forecast less room to improve.

\section{Discussion}
\label{sec:discussion}

The initial state of a frozen recurrent forecaster is a usable conditioning interface, and it
plays two roles worth keeping apart. As an adaptation surface, a static per-dataset state
recovers $8$--$70\%$ of the gain of full fine-tuning while training a fraction of the parameters.
As a channel for exogenous information, it does something fine-tuning cannot, since a
fine-tuned backbone still cannot reach a series' text without an architectural change to
accept a new modality. Whether a better conditioning signal could close the remaining gap
to full fine-tuning is open, since we tested only two.

Several scope conditions follow from how the study is set up. We treat TiRex as a case study
and do not claim the numbers carry over unchanged to other recurrent forecasters, though the
property we rely on, a recurrent state fixed to a default at inference, is shared across them.
Whether the improvements transfer is untested. Our backbone configuration is non-default, chosen as
described in Section~\ref{sec:setup} to make the block-type and decay effects
measurable.
We also leave a direct comparison with prompt and prefix tuning open.

Two fairly simple context vectors, a spectral summary of the window and an aligned text
embedding, are already enough to move a frozen foundation model, which suggests the interface
is worth future exploration. Richer context
vectors, other modalities and other recurrent backbones
all remain open. The initial state is an input that recurrent forecasters currently leave
idle, and we hope this opens an avenue others can build on.

\begin{ack}
This work was supported by the European Union's Horizon Europe research and innovation program under grant agreement number 101214398 (ELLIOT). The ELLIS Unit Linz, the LIT AI Lab, and the Institute for Machine Learning are supported by the Federal State of Upper Austria. NXAI GmbH provided all computing resources.
This work builds on the TiRex-2 codebase~\citep{podest2026tirex2generalizingtirexmultivariate}, developed by Patrick Podest, Marco Pichler, Elias B\"{u}rger, Levente Z\'{o}lyomi and Wilhelm Berghammer; we are particularly grateful to Marco Pichler and Patrick Podest for their support.
We thank Maximilian Beck and G\"{u}nter Klambauer for their feedback, David Stap for supporting the idea of a multimodal TiRex from which this work emerged, Andreas Auer for discussing such a model, and the anonymous reviewers for their comments.
B.V. thanks his mother, whose support throughout his studies made this work possible.
\end{ack}

{
\small

\bibliographystyle{unsrtnat}
\bibliography{references}
\par}
 
\clearpage
\appendix
\section*{Appendices}
\renewcommand{\thesection}{A\arabic{section}}
\renewcommand{\thetable}{A\arabic{table}}\setcounter{table}{0}
\renewcommand{\thefigure}{A\arabic{figure}}\setcounter{figure}{0}
\renewcommand{\theequation}{A\arabic{equation}}\setcounter{equation}{0}

Appendix~\ref{app:arch} specifies the backbone configuration and the state-initialisation
interface, including the parameter count of every head and control.
Appendix~\ref{app:fft} describes the spectral context vector,
Appendix~\ref{app:motime} the text pipeline and the separate CLIP alignment stage, and
Appendix~\ref{app:eval} the optimisation and evaluation settings shared by all runs.
Appendices~\ref{app:results}, \ref{app:static-decay}, \ref{app:gifteval} and
\ref{app:blocktype} give the per-dataset MoTime scores, the decay curves of the static
and text-conditioned states, the GiftEval scores by method and the per-seed localisation
scores. Appendix~\ref{app:shuffled} reports the shuffled-text control.

\section{Model architecture and state-initialisation interface}
\label{app:arch}

\begin{figure}[H]
\centering
\includegraphics[width=\linewidth,height=0.78\textheight,keepaspectratio]{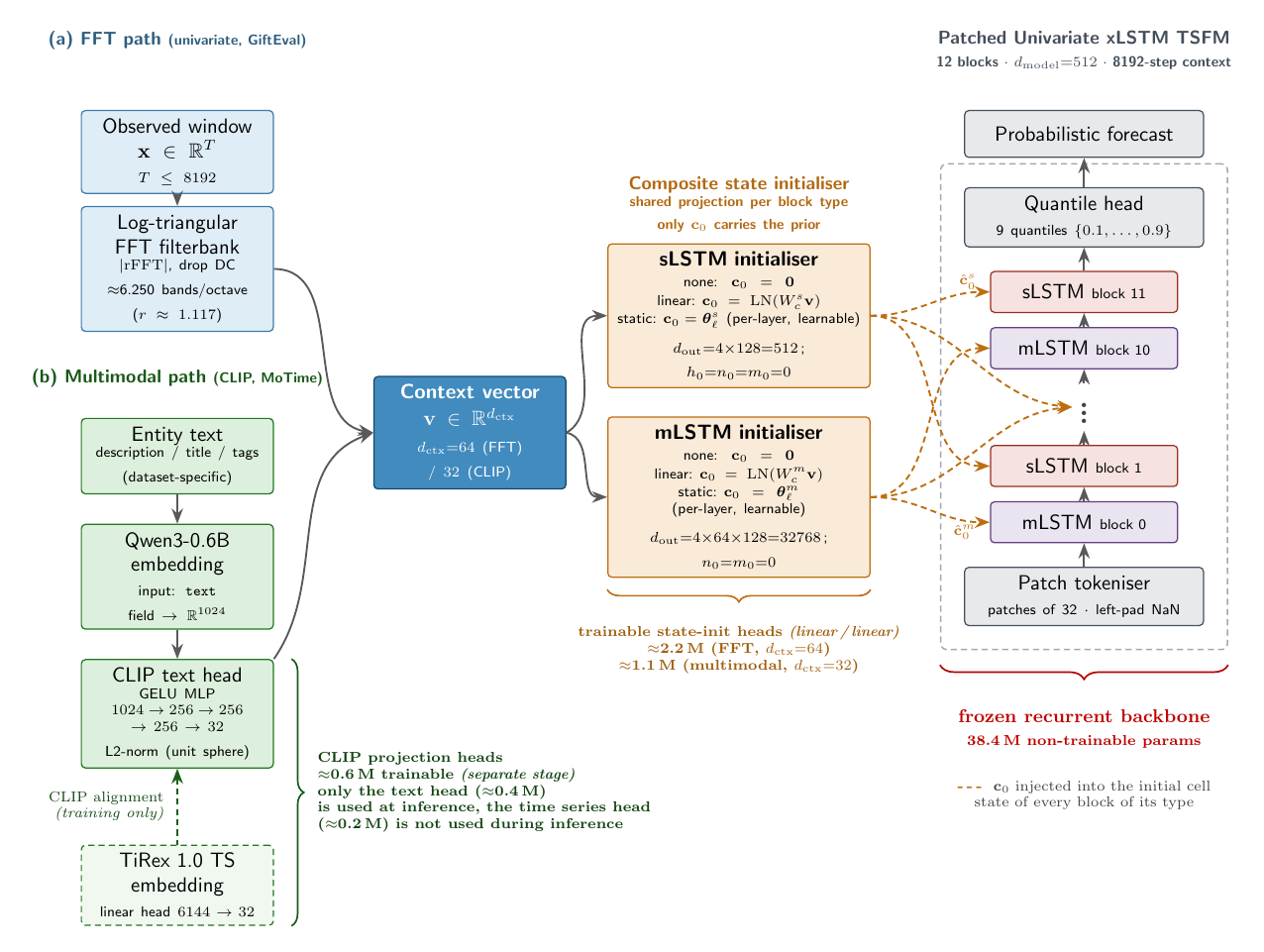}
\caption{Detailed architecture of the backbone and of the state-initialisation
  interface. The context vector is produced outside the backbone, either from the
  input window (spectral regime, Appendix~\ref{app:fft}) or from an aligned text
  embedding (exogenous regime, Appendix~\ref{app:motime}), and is written into the
  initial recurrent state of the xLSTM blocks.}
\label{fig:arch}
\end{figure}

\paragraph{Backbone.}
Table~\ref{tab:arch} lists the backbone configuration used in every experiment. It follows
the TiRex architecture with $12$ xLSTM blocks in a strictly alternating arrangement, so that
six blocks carry a scalar memory (sLSTM) and six a matrix memory (mLSTM), and the
contribution of each block type can be ablated separately. Figure~\ref{fig:arch} shows
the full data path, including the point at which the state-initialisation head writes
into the recurrent state.

\begin{table}[H]
\centering
\small
\caption{Backbone configuration, read from the training runs' own resolved configuration. The same configuration is used for every run in the paper; only the state initialiser and the set of trainable parameters change.}
\label{tab:arch}
\begin{tabular}{lp{0.62\linewidth}}
\toprule
Setting & Value \\
\midrule
Blocks & $12$, alternating; sLSTM at indices $1$, $3$, $5$, $7$, $9$, $11$ \\
Embedding dimension & $512$ \\
mLSTM heads & $4$; query/key dimension $256$, value dimension $512$ \\
sLSTM heads & $4$; head dimension $128$ \\
Input and output patch size & $32$, stride $32$ \\
Maximum context & $8{,}192$ steps \\
Quantile levels & $0.1$ to $0.9$ in steps of $0.1$ \\
Backbone parameters & $38.4$M, frozen in all state-init and LoRA runs \\
\bottomrule
\end{tabular}
\end{table}

\paragraph{What the head writes.}
Each xLSTM block is entered with a recurrent state that is normally all zero. In an
mLSTM block that state comprises a matrix memory
$C_0 \in \mathbb{R}^{4 \times 64 \times 128}$ ($\appxMlstmStateEntries$ entries per block), a
normaliser $n_0 \in \mathbb{R}^{4 \times 64}$ and a stabiliser
$m_0 \in \mathbb{R}^{4 \times 1}$. In an sLSTM block it comprises a cell state
$c_0 \in \mathbb{R}^{4 \times 128}$ ($\appxSlstmStateEntries$ entries per block) alongside the hidden,
normaliser and stabiliser slots. The head writes only the memory slot, $C_0$ in mLSTM
blocks and $c_0$ in sLSTM blocks; every other slot is left at zero. With
$c_0 = \mathbf{0}$ in every block the model reduces exactly to the released
zero-shot forecaster.

\paragraph{Head parameterisation.}
For a context vector of dimension $d$, the head applies one bias-free linear map
$\mathbb{R}^{d} \to \mathbb{R}^{D}$ followed by a LayerNorm over the $D$ outputs, with
$D = \appxSlstmStateEntries$ for sLSTM blocks and $D = \appxMlstmStateEntries$ for mLSTM blocks. One such map is shared
by all six blocks of a type, so the same vector is written into every block of that
type, and the parameter count is independent of the number of blocks. The static-state
control replaces the map with one learnable memory per block: a parameter tensor of
the same shape as that block's memory slot, which receives no input. It is
initialised at zero, so training starts from the zero-shot model, and is non-zero
once trained; what it learns is a single fixed state written into every forecast,
identical for every series. Table~\ref{tab:head_params}
gives the resulting counts.

\begin{table}[H]
\centering
\small
\caption{Trainable parameters of each state-initialisation variant, counted by instantiating the initialisers and configuring them exactly as a run does. Linear terms are $d \cdot D$ and LayerNorm terms $2D$, with $D = 512$ (sLSTM) and $D = 32{,}768$ (mLSTM). The static state receives no input and stores one learnable memory per block, so it scales with the block count rather than with $d$.}
\label{tab:head_params}
\begin{tabular}{lrrrr}
\toprule
Variant & $d$ & sLSTM part & mLSTM part & Total \\
\midrule
Spectral head (GiftEval) & $64$ & $33{,}792$ & $2{,}162{,}688$ & $2{,}196{,}480$ \\
Text head (MoTime) & $32$ & $17{,}408$ & $1{,}114{,}112$ & $1{,}131{,}520$ \\
Static state & -- & $3{,}072$ & $196{,}608$ & $199{,}680$ \\
\bottomrule
\end{tabular}
\end{table}

\paragraph{LoRA control.}
The LoRA control adds a rank-$r$ update $\Delta W = \tfrac{\alpha}{r} B A$ to selected
linear layers inside the block stack, with $\alpha = r$, $A$ initialised
Kaiming-uniform and $B$ initialised to zero, so an injected model is numerically
identical to the frozen one before training. Adapted layers are the five mLSTM
projections (query, key, value, output gate pre-activation, output projection) in each
of the six mLSTM blocks, and the three feed-forward projections (gate, up, down) in all
twelve blocks. The patch embedding and the output head are excluded, following the
common practice of leaving embeddings and output heads unadapted. The sLSTM cell itself
is not adapted, because its gates are block-diagonal head-wise
expansions and its recurrent kernel is a fused CUDA extension rather than a set of
adaptable linear layers; sLSTM blocks are therefore
adapted only through their feed-forward projections. On this geometry each rank adds $\appxLoraPerRank$ parameters, which gives the counts in Table~\ref{tab:lora_params} and lets
each rank be placed just above the head it is matched against.

\begin{table}[H]
\centering
\small
\caption{LoRA parameter counts on this backbone geometry and the head each rank is matched to. The per-rank cost of $96{,}768$ parameters is measured by injecting the adapter into the 66 target layers with the same routine the runs use.}
\label{tab:lora_params}
\begin{tabular}{lrl}
\toprule
Rank & Parameters & Matched against \\
\midrule
$12$ & $1{,}161{,}216$ & text head, $1{,}131{,}520$ \\
$23$ & $2{,}225{,}664$ & spectral head, $2{,}196{,}480$ \\
$24$ & $2{,}322{,}432$ & LoRA $r{=}12$ plus text head, $2{,}292{,}736$ \\
$46$ & $4{,}451{,}328$ & doubled-budget control for rank $23$ \\
\bottomrule
\end{tabular}
\end{table}

\section{Spectral conditioning details}
\label{app:fft}

In the spectral regime the context vector is a fixed-size summary of the input window
itself, so no data beyond the window is used. The same creator object produces the
vector during training and during evaluation, which guarantees that the map from window
to $c_0$ is identical in both phases.

\paragraph{From window to context vector.}
The pipeline has four stages.
\emph{(i) Length alignment.} The window is aligned to a fixed length of $8{,}192$
steps: longer windows are truncated to their most recent $8{,}192$ observations, shorter
windows are padded on the left with missing values. The context vector therefore has the
same dimension for every series, at the cost of a padded prefix for short windows.
\emph{(ii) Missing-value filling.} A leading run of missing values is set to zero, an
interior gap is filled by linear interpolation between its neighbours, and a trailing
gap decays linearly to zero. A window with no valid observation yields an all-zero
series.
\emph{(iii) Spectrum.} The window mean is subtracted and the magnitude of the real FFT
is taken. The DC bin is dropped, leaving bins $1$ to $4{,}096$.
\emph{(iv) Filterbank.} The magnitude spectrum is projected onto a bank of
logarithmically spaced triangular filters, one output feature per filter.

\paragraph{Filterbank construction.}
Centre bins are the rounded values of a continuous geometric sequence over bin index,
$b_j = \lfloor s\,r^{\,j} + 0.5 \rfloor$ with start bin $s = 1$ and ratio $r = 2^{1/B}$,
taken for every $j$ with $s\,r^{\,j}$ at most the highest available bin and clamped to that
range. Rounding maps several consecutive $j$ onto the same bin at low frequencies, and those
duplicates are removed; the highest available bin is appended as the final centre if the
sequence has not already reached it.
Writing the surviving centres as $c_1 < c_2 < \dots < c_M$, filter $i$ is a triangle that
rises linearly from $c_{i-1}$ to $c_i$ and falls linearly to $c_{i+1}$, with a weight of
exactly $1$ at $c_i$; the first and last filters are the corresponding half-triangles. Filters are not area-normalised, and the resulting
vector is used without further normalisation and without appended summary statistics.
We set $B = 6.24985564964451$ bands per octave, that is $r = 1.11729$, which is the
value that makes the bank contain exactly $64$ filters at an $8{,}192$-step window and
so fixes the context dimension at $d = 64$. The resulting centre bins are
\appxFftCentreBins. Resolution is therefore one filter per bin at the
lowest frequencies and progressively coarser towards the Nyquist bin.

\paragraph{Injection and ablation.}
The $64$-dimensional vector is written into the memory slot of every block through the
shared heads of Table~\ref{tab:head_params}, for a total of $\appxSpectralHeadParams$ trainable parameters. The localisation ablation of Appendix~\ref{app:blocktype} replaces one of
the two heads by an initialiser that writes nothing, so the parameters of the disabled
side are absent rather than merely unused: the mLSTM-only arm trains $\appxSpectralHeadMlstmParams$ parameters and the sLSTM-only arm $\appxSpectralHeadSlstmParams$. The two arms are consequently not
parameter-matched to each other, since the state sizes they write into differ by a factor of $\appxStateSizeRatio$.

\paragraph{Controls in this regime.}
The static-state control writes a learned per-block constant and receives no context
vector, which controls for the value of a non-zero learnable starting point. The LoRA controls receive no context vector either and adapt
the backbone instead. In the joint arm the spectral head and a rank-$23$ adapter are
trained together, against a rank-$46$ adapter of comparable total size.
 
\section{Exogenous (text) conditioning details}
\label{app:motime}

In the exogenous regime the context vector is derived from text associated with each
series, which the numeric window does not contain. The pipeline has three stages that
are run once per dataset and then frozen: dataset construction, embedding, and CLIP
alignment. Only the state-initialisation head is trained inside the forecasting run.

\paragraph{Dataset construction.}
Each MoTime source is converted into a daily-frequency dataset of univariate series
with one associated text per series. Series are split at the series level into training,
validation and test partitions in the ratio $80/10/10$ with seed $42$, so the partitions
share no series and the head must map texts it has not seen at evaluation time. The
missing-value sentinel is mapped to missing, and series with fewer than eight
non-missing observations are dropped; this affects only News-FO, where $235$ of $26{,}612$
series are removed. For Wiki People, where several traffic channels exist per entity, the
all-access, all-agents channel is used. Table~\ref{tab:motime_data} reports the resulting
sizes together with the evaluation task parameters.

\begin{table}[H]
\centering
\small
\caption{MoTime datasets after conversion, and the evaluation task per dataset. Series are split at the series level with seed $42$ in the ratio $80/10/10$, and series with fewer than $8$ non-missing observations are dropped. All tasks use horizon $32$, a minimum context of $32$ observations and $9$ quantile levels; windows are rolling with the given step. Evaluation is on the validation partition. News-FO has variable series lengths.}
\label{tab:motime_data}
\resizebox{\linewidth}{!}{%
\begin{tabular}{lrrrrrrrr}
\toprule
Dataset & Series length & Series kept & Train & Validation & Test & Windows & Window step & Seasonality \\
\midrule
MovieLens & $10{,}505$ & $84{,}432$ & $67{,}546$ & $8{,}443$ & $8{,}443$ & $28$ & $365$ & $7$ \\
News-FO & variable & $26{,}377$ & $21{,}101$ & $2{,}638$ & $2{,}638$ & $3$ & $32$ & $1$ \\
PixelRec Full & $4{,}865$ & $43{,}082$ & $34{,}466$ & $4{,}308$ & $4{,}308$ & $20$ & $224$ & $7$ \\
PixelRec Sub & $1{,}308$ & $998$ & $798$ & $100$ & $100$ & $20$ & $64$ & $7$ \\
Tianchi & $184$ & $36{,}397$ & $29{,}117$ & $3{,}640$ & $3{,}640$ & $4$ & $32$ & $7$ \\
Wiki People & $550$ & $918$ & $734$ & $92$ & $92$ & $16$ & $32$ & $7$ \\
\bottomrule
\end{tabular}}
\end{table}

\paragraph{Embedding.}
The text of each series is embedded once with Qwen3-Embedding-0.6B into
$\appxTextEmbedDim$ dimensions. The series itself is embedded once with the publicly
released TiRex 1.0 model: its per-layer hidden states are mean-pooled over patches,
$L^2$-normalised per layer and concatenated across layers, giving
$\appxSeriesEmbedDim$ dimensions. Both embeddings are stored with the dataset and are
never recomputed.

\paragraph{CLIP alignment.}
The two embedding spaces are aligned per dataset with a CLIP-style contrastive stage.
The text branch is a multi-layer perceptron with three hidden layers of width $256$ and
GELU activations, mapping $\appxClipTextWidths$. The series branch is
a single linear map $\appxClipSeriesWidths$. Both outputs are $L^2$-normalised, giving
$\appxClipTotalParams$ trainable parameters in total ($\appxClipTextParams$ in the text branch and $\appxClipSeriesParams$ in the series branch). Given a batch of paired embeddings, the loss is the symmetric
InfoNCE objective: cosine similarities are divided by a temperature of $0.07$ and cross
entropy against the diagonal is averaged over both directions. Optimisation uses AdamW
at a constant learning rate of $10^{-5}$ with batch size $4{,}096$ and seed $42$; no
learning-rate schedule is applied. The stage is fitted on the training partition only.
Ten percent of that partition is held out as an internal validation split, and training
stops when the contrastive validation loss fails to improve by at least $10^{-4}$ for
$100$ consecutive epochs, under an epoch cap of $100{,}000$; the parameters at the best
validation loss are retained. Raw embeddings are passed through without pre-normalisation
and without norm-percentile clipping. Table~\ref{tab:clip_loss} reports the contrastive
validation loss before and after this stage.

\begin{table}[H]
\centering
\small
\caption{Symmetric InfoNCE loss on the internal validation split of each dataset's training partition, at initialisation and at the selected early-stopping point (patience $100$ epochs, minimum improvement $10^{-4}$). Lower is better.}
\label{tab:clip_loss}
\begin{tabular}{lrrr}
\toprule
Dataset & Initial & Best & Reduction \\
\midrule
MovieLens & $9.3996$ & $7.7955$ & $17.1\%$ \\
News-FO & $8.2661$ & $7.3685$ & $10.9\%$ \\
PixelRec Full & $9.1939$ & $7.8305$ & $14.8\%$ \\
PixelRec Sub & $5.5657$ & $4.3007$ & $22.7\%$ \\
Tianchi & $8.9369$ & $6.9089$ & $22.7\%$ \\
Wiki People & $4.9982$ & $3.8171$ & $23.6\%$ \\
\bottomrule
\end{tabular}
\end{table}

\paragraph{Injection.}
The aligned $\appxSharedDim$-dimensional text projection is precomputed for every series in every
partition and stored as a dataset column. During a forecasting run the collator reads
that column and passes the vector to the shared heads of Table~\ref{tab:head_params},
for $\appxTextHeadParams$ trainable parameters. The alignment stage is frozen at this point,
so its $\appxClipTotalParams$ parameters are not part of the trainable budget and are not counted
in the comparisons against the LoRA controls. The static-state control consumes no
column, and the LoRA controls consume no column and adapt the backbone instead. Both
therefore have no access to the text.
 
\section{Training and evaluation settings}
\label{app:eval}

All runs start from the same pretrained checkpoint of the configuration in
Table~\ref{tab:arch} and share the optimisation settings in Table~\ref{tab:optim}. What differs between arms is the set of
parameters that receive gradients (Table~\ref{tab:trainable}), the training corpus, and
the step budget.

\begin{table}[H]
\centering
\small
\caption{Optimisation settings shared by every run. The optimiser base learning rate is $0$ and the schedule sets the effective rate, so $2 \times 10^{-4}$ is the peak value reached after the warm-up fraction.}
\label{tab:optim}
\begin{tabular}{lp{0.62\linewidth}}
\toprule
Setting & Value \\
\midrule
Optimiser & AdamW, weight decay $0.01$ \\
Schedule & one-cycle, peak learning rate $2 \times 10^{-4}$, warm-up fraction $0.05$, no momentum cycling \\
Precision & bfloat16 mixed \\
Batch size & $256$, no gradient accumulation \\
Loss & quantile loss over the nine levels $0.1$ to $0.9$, label offset of one output patch, no per-sample impact cap \\
Hardware & one H100 per run \\
\bottomrule
\end{tabular}
\end{table}

\begin{table}[H]
\centering
\small
\caption{Parameter groups that receive gradients in each arm. In every arm except full
  fine-tuning the backbone weights stay at their pretrained values.}
\label{tab:trainable}
\begin{tabular}{lll}
\toprule
Arm & Trainable parameters & Backbone \\
\midrule
State-init head, static state & state initialiser only & frozen \\
LoRA & adapter factors only & frozen \\
LoRA and state init & adapter factors and state initialiser & frozen \\
Full fine-tuning & all parameters & updated \\
Zero-shot & none, evaluation only & frozen \\
\bottomrule
\end{tabular}
\end{table}

\paragraph{Spectral runs on GiftEval.}
The spectral head and its controls are trained for $\appxGiftevalSteps$ steps at batch size
$256$, which is $\appxGiftevalSamplesM$M training samples per run. Training uses the TiRex 1.1 pretraining recipe, data mix and
augmentation stack unchanged, so the head is fitted
only on data the backbone has already seen.
Missing values are left in place rather than imputed, each series is truncated to its
most recent $8{,}192$ observations, and the next output patch is extracted as the label. GiftEval is scored at the final step of each run. Six seeds
per arm are used, $1$ through $6$, controlling head initialisation and data order. The
spectral regime therefore comprises $30$ trained runs over its five arms (spectral head,
static state, LoRA $r{=}23$, LoRA $r{=}46$, and the joint spectral-and-LoRA arm), a
further $12$ trained runs for the two single-block arms of
Appendix~\ref{app:blocktype}, and one zero-shot evaluation.

\paragraph{Exogenous runs on MoTime.}
One run is trained per combination of dataset, method and seed, with seeds
$\{7, 11, 42, 123, 456, 789\}$. Six methods over six datasets and six seeds gives $216$
trained runs, plus one zero-shot evaluation per dataset. The training corpus is the
training partition of the dataset itself, streamed rather than iterated in epochs, with the same
augmentation stack. Step budgets are set per dataset in
proportion to the size of its training partition, at approximately ten passes over that
partition, and are held fixed across all six methods so that budget is not a confounder
between them. Table~\ref{tab:motime_steps} gives the exact values. Every run is
scored on the dataset's evaluation task at its final step, and that score is the one
reported.

\begin{table}[H]
\centering
\small
\caption{Per-dataset step budgets for the MoTime runs, identical across the six methods. Samples drawn is the budget times the batch size of $256$; passes is that figure divided by the number of training series. Training streams from the dataset, so no epoch boundary exists and the pass count is nominal.}
\label{tab:motime_steps}
\begin{tabular}{lrrrr}
\toprule
Dataset & Train series & Steps & Samples drawn & Passes \\
\midrule
MovieLens & $67{,}546$ & $2{,}640$ & $675{,}840$ & $10.0$ \\
News-FO & $21{,}101$ & $830$ & $212{,}480$ & $10.1$ \\
PixelRec Full & $34{,}466$ & $1{,}350$ & $345{,}600$ & $10.0$ \\
PixelRec Sub & $798$ & $40$ & $10{,}240$ & $12.8$ \\
Tianchi & $29{,}117$ & $1{,}140$ & $291{,}840$ & $10.0$ \\
Wiki People & $734$ & $30$ & $7{,}680$ & $10.5$ \\
\bottomrule
\end{tabular}
\end{table}

\paragraph{Evaluation.}
GiftEval is evaluated through its benchmark harness over the full set of tasks, and we
report the aggregate MASE and the aggregate scaled quantile loss, which that harness
labels CRPS. MoTime is evaluated through the same harness with the per-dataset task
definitions of Table~\ref{tab:motime_data}; MASE is reported in the paper, and the
scaled quantile loss and RMSE are computed alongside it. Every number reported in the
paper is obtained without test-time augmentation. Test-time augmentation is applied only inside the prediction path, which is separate from the training step, so
the choice of postprocessor affects the evaluation and not the trained parameters.

\section{GiftEval results by method}
\label{app:gifteval}
Aggregate GiftEval scores behind Figure~\ref{fig:gifteval} (left): the spectral
state against every control, on the pretrained checkpoint without test-time
augmentation. All five trained arms share the corpus, the step budget, the
optimisation settings of Table~\ref{tab:optim} and six seeds ($1$ through $6$);
they differ only in what receives gradients and in whether a context vector is
written into $c_0$.

\begin{table}[H]
\centering
\caption{GiftEval aggregate scores for FFT state initialisation and its
  capacity-matched controls, on the pretrained checkpoint without test-time
  augmentation. Mean $\pm$ sample standard deviation over seeds; $\Delta$ is
  relative to the zero-shot ($\mathbf{c}_0=\mathbf{0}$) baseline. Lower is better.}
\label{tab:fft_gifteval_base_notta}
\resizebox{\linewidth}{!}{%
\begin{tabular}{lcrrrr}
\toprule
Method & $n$ & MASE & $\Delta$ & CRPS & $\Delta$ \\
\midrule
FFT init & 6 & $0.70469 \pm 0.00037$ & $-0.175\%$ & $0.47995 \pm 0.00027$ & $-0.206\%$ \\
static prior & 6 & $0.70606 \pm 0.00014$ & $+0.019\%$ & $0.48104 \pm 0.00010$ & $+0.021\%$ \\
LoRA r23 & 6 & $0.70625 \pm 0.00020$ & $+0.046\%$ & $0.48113 \pm 0.00013$ & $+0.039\%$ \\
LoRA r23 + FFT & 6 & $0.72109 \pm 0.00088$ & $+2.147\%$ & $0.48942 \pm 0.00057$ & $+1.762\%$ \\
LoRA r46 & 6 & $0.70618 \pm 0.00032$ & $+0.036\%$ & $0.48103 \pm 0.00020$ & $+0.019\%$ \\
\midrule
Zero-shot baseline ($\mathbf{c}_0=\mathbf{0}$) & 1 & $0.70593$ & -- & $0.48094$ & -- \\
\bottomrule
\end{tabular}}
\end{table}

The three controls isolate different explanations for the spectral head's gain.
The static state has the same expressivity, one learned tensor per block, but no
input, so it controls for the value of a non-zero starting point. LoRA $r{=}23$ spends a matched trainable budget on the
weights instead of the state. LoRA $r{=}46$ doubles that budget, so a reading of
the effect as "more trainable parameters help" would predict it to do better than
LoRA $r{=}23$. The joint arm trains the spectral head and a rank-$23$ adapter
together, against LoRA $r{=}46$ as its budget-matched counterpart.

Per-seed scores for the injection-locus arms are in
Appendix~\ref{app:blocktype}; the \emph{both} arm there is the same six runs as
FFT init in Table~\ref{tab:fft_gifteval_base_notta}.

\section{Block-type localisation}
\label{app:blocktype}
Per-seed GiftEval scores behind Figure~\ref{fig:gifteval} (right), on the pretrained
checkpoint without test-time augmentation.
 
\begin{table}[H]
\centering
\caption{GiftEval aggregate scores for FFT state initialisation by injection
  locus, on the pretrained checkpoint without test-time augmentation. Mean $\pm$
  sample standard deviation over 6 seeds; $\Delta$ is relative to the zero-shot
  ($\mathbf{c}_0=\mathbf{0}$) baseline. Lower is better.}
\label{tab:fft_blocktype_base_notta}
\resizebox{\linewidth}{!}{%
\begin{tabular}{lcrrrr}
\toprule
Injection locus & $n$ & MASE & $\Delta$ & CRPS & $\Delta$ \\
\midrule
both & 6 & $0.70469 \pm 0.00037$ & $-0.175\%$ & $0.47995 \pm 0.00027$ & $-0.206\%$ \\
mLSTM only & 6 & $0.70511 \pm 0.00016$ & $-0.116\%$ & $0.47974 \pm 0.00010$ & $-0.249\%$ \\
sLSTM only & 6 & $0.70694 \pm 0.00017$ & $+0.144\%$ & $0.48205 \pm 0.00012$ & $+0.230\%$ \\
\midrule
Zero-shot baseline ($\mathbf{c}_0=\mathbf{0}$) & 1 & $0.70593$ & -- & $0.48094$ & -- \\
\bottomrule
\end{tabular}}
\end{table}

The three arms differ only in where the spectral vector is written. All three train the
same spectral head of Appendix~\ref{app:fft} on the same corpus
for $\appxGiftevalSteps$ steps under the settings of
Table~\ref{tab:optim}, with six seeds each. In the single-block arms the head for the
disabled block type is replaced by an initialiser that writes nothing, so that side
contributes no parameters: the mLSTM-only arm trains $\appxSpectralHeadMlstmParams$ parameters, the sLSTM-only arm $\appxSpectralHeadSlstmParams$, and the arm that writes into both $\appxSpectralHeadParams$. The arms
are therefore not parameter-matched to each other, because the memory slot of an mLSTM
block holds $\appxMlstmStateEntries$ entries against $\appxSlstmStateEntries$ in an sLSTM block. Per-seed scores are given
in Table~\ref{tab:fft_blocktype_base_notta_per_seed}.

\begin{table}[H]
\centering
\caption{Per-seed GiftEval aggregate scores behind
  Table~\ref{tab:fft_blocktype_base_notta} (pretrained checkpoint, no test-time augmentation).}
\label{tab:fft_blocktype_base_notta_per_seed}
\resizebox{\linewidth}{!}{%
\begin{tabular}{lrrrrrrrr}
\toprule
Injection locus & $s{=}1$ & $s{=}2$ & $s{=}3$ & $s{=}4$ & $s{=}5$ & $s{=}6$ & mean & std \\
\midrule
\multicolumn{9}{l}{\textit{MASE}} \\
\quad both & $0.70454$ & $0.70504$ & $0.70501$ & $0.70494$ & $0.70412$ & $0.70450$ & $0.70469$ & $0.00037$ \\
\quad mLSTM only & $0.70498$ & $0.70522$ & $0.70510$ & $0.70535$ & $0.70513$ & $0.70490$ & $0.70511$ & $0.00016$ \\
\quad sLSTM only & $0.70676$ & $0.70683$ & $0.70685$ & $0.70716$ & $0.70716$ & $0.70690$ & $0.70694$ & $0.00017$ \\
\quad Zero-shot baseline & -- & -- & -- & -- & -- & -- & $0.70593$ & -- \\
\multicolumn{9}{l}{\textit{CRPS}} \\
\quad both & $0.47981$ & $0.48018$ & $0.48006$ & $0.48025$ & $0.47952$ & $0.47989$ & $0.47995$ & $0.00027$ \\
\quad mLSTM only & $0.47969$ & $0.47978$ & $0.47980$ & $0.47987$ & $0.47976$ & $0.47957$ & $0.47974$ & $0.00010$ \\
\quad sLSTM only & $0.48192$ & $0.48194$ & $0.48216$ & $0.48204$ & $0.48201$ & $0.48221$ & $0.48205$ & $0.00012$ \\
\quad Zero-shot baseline & -- & -- & -- & -- & -- & -- & $0.48094$ & -- \\
\bottomrule
\end{tabular}}
\end{table}

\section{Per-dataset results MoTime}
\label{app:results}
Per-method, per-dataset MASE behind Figure~\ref{fig:motime_methods}.
 
\begin{table}[H]
\centering
\small
\caption{MoTime MASE per method and dataset (lower is better). Mean $\pm$ sample
  standard deviation over seeds; $n$ is the number of seeds. The zero-shot baseline
  (frozen backbone, $\mathbf{c}_0=\mathbf{0}$) is a single run per dataset.}
\label{tab:motime_methods_mase}
\resizebox{\linewidth}{!}{%
\begin{tabular}{lcrrrrrr}
\toprule
Method & $n$ & MovieLens & News-FO & PixelRec Full & PixelRec Sub & Tianchi & Wiki People \\
\midrule
static & 6 & $2.32794 \pm 0.0002$ & $0.363826 \pm 0.0003$ & $13.2940 \pm 0.0015$ & $5.50899 \pm 0.00074$ & $48.3219 \pm 0.0015$ & $3.95267 \pm 0.00077$ \\
text & 6 & $2.33438 \pm 0.0013$ & $0.365138 \pm 0.00079$ & $13.2400 \pm 0.012$ & $5.49286 \pm 0.0023$ & $48.2499 \pm 0.007$ & $3.94056 \pm 0.0021$ \\
LoRA r12 & 6 & $2.31419 \pm 0.0016$ & $0.353915 \pm 0.00015$ & $13.2119 \pm 0.0024$ & $5.50333 \pm 0.00064$ & $48.2222 \pm 0.0045$ & $3.94647 \pm 0.00058$ \\
LoRA+text & 6 & $2.31241 \pm 0.0021$ & $0.356787 \pm 0.00072$ & $13.1910 \pm 0.0045$ & $5.48712 \pm 0.0021$ & $48.1775 \pm 0.0073$ & $3.93716 \pm 0.0019$ \\
LoRA r24 & 6 & $2.31160 \pm 0.0025$ & $0.352768 \pm 0.00017$ & $13.2167 \pm 0.002$ & $5.49763 \pm 0.00052$ & $48.2051 \pm 0.0056$ & $3.93968 \pm 0.0013$ \\
full FT & 6 & $2.27991 \pm 0.0044$ & $0.351926 \pm 0.0002$ & $13.2190 \pm 0.0083$ & $5.47458 \pm 0.0016$ & $48.1502 \pm 0.0068$ & $3.92771 \pm 0.0016$ \\
\midrule
zero-shot ($\mathbf{c}_0=\mathbf{0}$) & 1 & $2.33189$ & $0.391027$ & $13.3525$ & $5.51275$ & $48.3407$ & $3.95633$ \\
\bottomrule
\end{tabular}}
\end{table}

\section{Static and text-conditioned state decay}
\label{app:static-decay}

\paragraph{Method.}
The decay measurement isolates the effect of the initial state by holding the forecast
target fixed and varying only how much history the model is allowed to see. It is
anchored on the evaluation windows of Table~\ref{tab:motime_data}: for a task with
horizon $H$, $W$ windows and step $S$, window $w$ has its cutoff at $-H - (W - 1 - w)S$
observations from the end of the series, and a window participates if the history
available before its cutoff reaches the task minimum of $32$ observations. For each
window the visible history is then truncated to $k$ patches, that is $32k$ observations
with the most recent kept, for every $k$ from $1$ up to that window's full history and
capped at the model's $8{,}192$-step limit. Every resulting instance is forecast twice on
the same frozen backbone, once from $c_0 = \mathbf{0}$ and once with the trained state
initialiser, and the two forecasts are scored against the same target. Whether the text
column is read at prediction time follows from the initialiser, so the static-state runs
consume none.

MASE and the scaled quantile loss are computed with the harness's own components, but
the seasonal error is taken from the full history up to the cutoff and held fixed across
all truncations of that instance, so that values are comparable along the length axis.
Following the harness, an undefined seasonal error is replaced by one. Instances whose
full history is constant have a seasonal error of zero and are dropped, as are grid
points with fewer than $100$ surviving paired instances. Each dataset contributes a
deterministic subsample of $500$ series. The reported quantity is the mean change in MASE over paired instances, trained initialisation minus zero initialisation, so a negative value means the initialiser lowers the error at that context length. The mean is used
rather than the median because the training objective is the quantile loss at level
$0.5$, whose numerator is the same absolute error that MASE averages.

Table~\ref{tab:decay_grid} gives the grid extent per dataset. The extent is set by the
series lengths, so Tianchi and News-FO admit only a few grid points while MovieLens
reaches the model's context limit. The source runs are the single-seed text-conditioned
and static-state runs at seed $11$.

\begin{table}[H]
\centering
\small
\caption{Grid extent of the decay measurement. Paired instances are reported at the shortest and the longest grid point; the count falls with context length because fewer windows have that much history available. Where two counts are listed, the text-conditioned and static-state runs do not rest on the same instance set at that grid point.}
\label{tab:decay_grid}
\resizebox{\linewidth}{!}{%
\begin{tabular}{lrrrrr}
\toprule
Dataset & Grid points & Context range & Windows & Instances at the shortest point & Instances at the longest point \\
\midrule
MovieLens & $256$ & $32$ to $8{,}192$ & $28$ & $4{,}060$ & $2{,}662$ \\
News-FO & $3$ & $32$ to $96$ & $3$ & $1{,}305$ & $438$ \\
PixelRec Full & $151$ & $32$ to $4{,}832$ & $20$ & $1{,}833$ & $500$ \\
PixelRec Sub & $39$ & $32$ to $1{,}248$ & $20$ & $1{,}770$ & $100$ \\
Tianchi & $4$ & $32$ to $128$ & $4$ & $909$ / $1{,}584$ & $470$ \\
Wiki People & $15$ & $32$ to $480$ & $16$ & $1{,}472$ & $184$ \\
\bottomrule
\end{tabular}}
\end{table}

For every dataset except Tianchi the two variants are measured on the same instance set.
For Tianchi the text-conditioned run covers $4{,}600$ per-instance rows over $515$ series
and the static-state run $7{,}000$ rows over $500$ series, so its two shortest grid points
rest on different instance sets ($909$ against $1{,}584$ at $32$ observations, and $899$
against $1{,}264$ at $64$); the two longest grid points agree at $885$ and $470$. The
Tianchi panels are therefore not paired between variants at short context.

\begin{figure}[H]
\centering
\begin{minipage}{0.49\linewidth}\centering
\includegraphics[width=\linewidth]{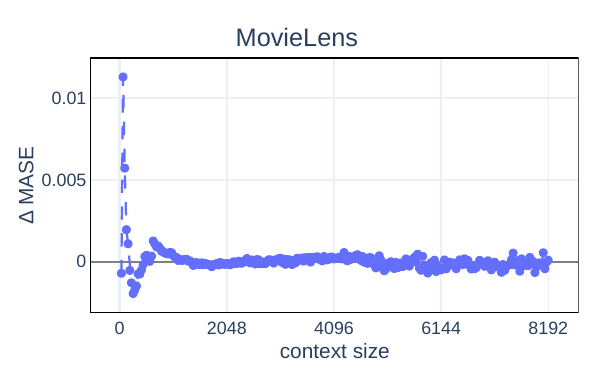}
\end{minipage}\hfill
\begin{minipage}{0.49\linewidth}\centering
\includegraphics[width=\linewidth]{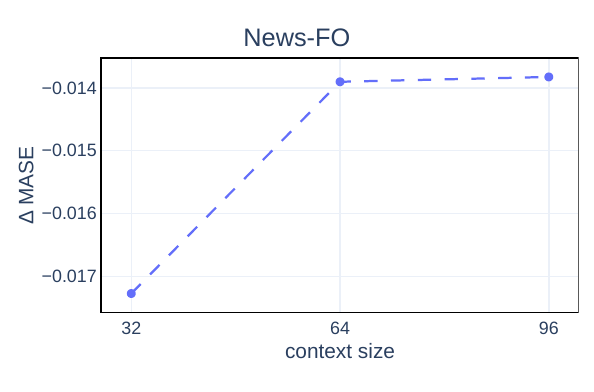}
\end{minipage}

\begin{minipage}{0.49\linewidth}\centering
\includegraphics[width=\linewidth]{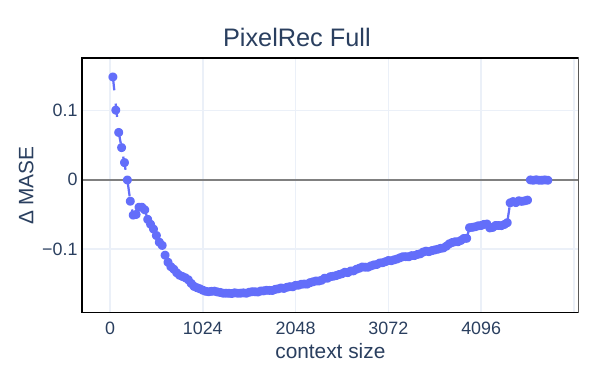}
\end{minipage}\hfill
\begin{minipage}{0.49\linewidth}\centering
\includegraphics[width=\linewidth]{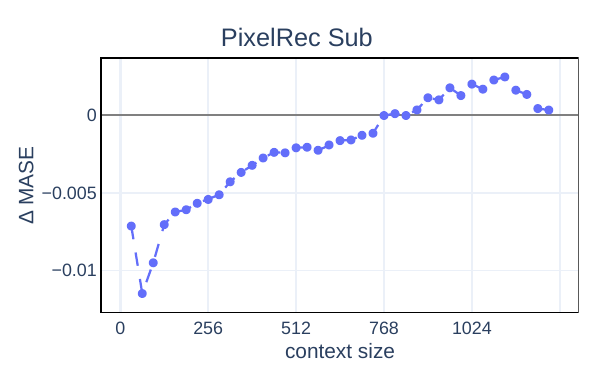}
\end{minipage}

\begin{minipage}{0.49\linewidth}\centering
\includegraphics[width=\linewidth]{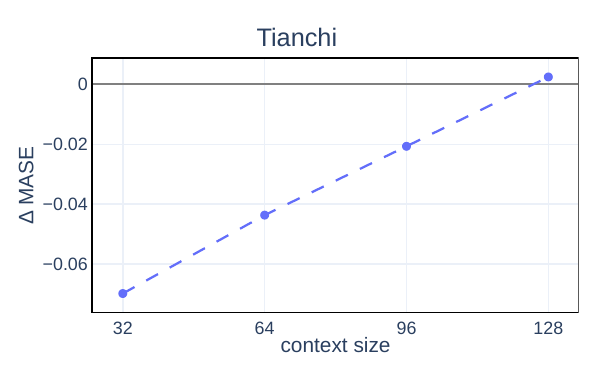}
\end{minipage}\hfill
\begin{minipage}{0.49\linewidth}\centering
\includegraphics[width=\linewidth]{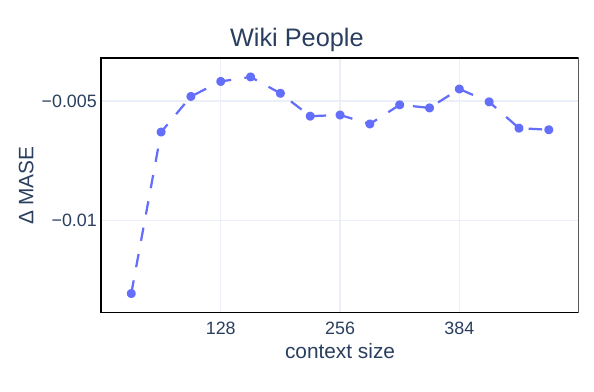}
\end{minipage}
\caption{Static state, mean change in MASE caused by the injected state (with minus without; negative means the initialiser lowers the error) against the number of observations shown. Grid extents differ by dataset as given in
  Table~\ref{tab:decay_grid}.}
\label{fig:decay_static_app}
\end{figure}

\begin{figure}[H]
\centering
\begin{minipage}{0.49\linewidth}\centering
\includegraphics[width=\linewidth]{resources/export/state_decay/movielens_fullset_linear_mase_diff_mean_with_title.pdf}
\end{minipage}\hfill
\begin{minipage}{0.49\linewidth}\centering
\includegraphics[width=\linewidth]{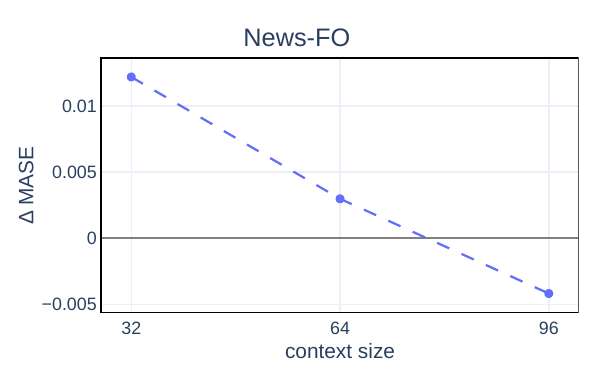}
\end{minipage}

\begin{minipage}{0.49\linewidth}\centering
\includegraphics[width=\linewidth]{resources/export/state_decay/pixelrec_fullset_linear_mase_diff_mean_with_title.pdf}
\end{minipage}\hfill
\begin{minipage}{0.49\linewidth}\centering
\includegraphics[width=\linewidth]{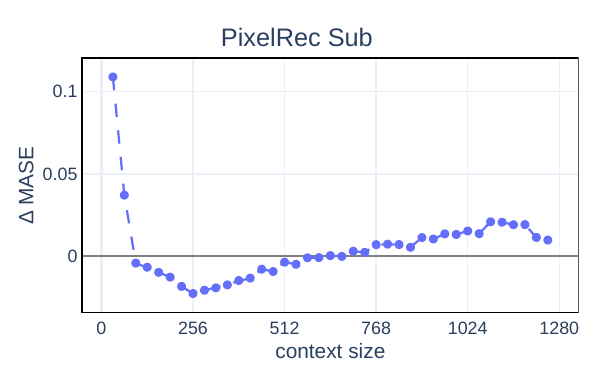}
\end{minipage}

\begin{minipage}{0.49\linewidth}\centering
\includegraphics[width=\linewidth]{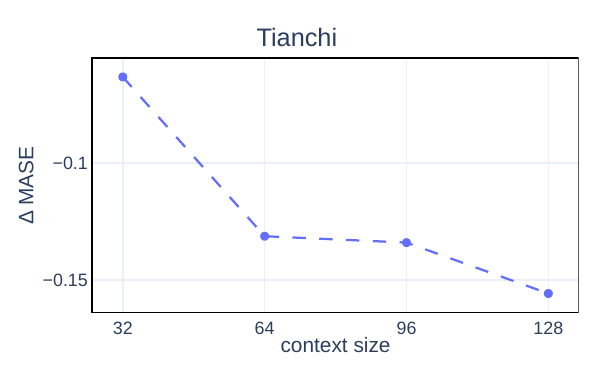}
\end{minipage}\hfill
\begin{minipage}{0.49\linewidth}\centering
\includegraphics[width=\linewidth]{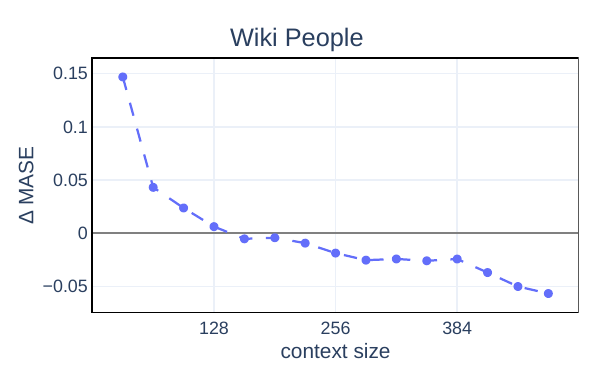}
\end{minipage}
\caption{Text-conditioned state, mean change in MASE caused by the injected state (with minus without; negative means the initialiser lowers the error) against the number of observations shown. The MovieLens and PixelRec Full panels are the data of
  Figure~\ref{fig:decay}.}
\label{fig:decay_linear_app}
\end{figure}

Figure~\ref{fig:decay_static_app} shows the static-state curves for all six datasets and
Figure~\ref{fig:decay_linear_app} the text-conditioned curves, including the two datasets
shown in Figure~\ref{fig:decay}.

\section{Shuffled-text control}
\label{app:shuffled}

To test whether the gain of the text-conditioned state depends on the text belonging to
its series, we repeat the text and LoRA+text runs with every series paired with the text
of another series. Within each partition, the text, its Qwen3 embedding and its CLIP
projection are reassigned by a fixed random derangement, so that no series keeps its own
text; the series, the partitions and all training settings are unchanged, and the
shuffled models are trained and evaluated on this assignment. As a second control, which
requires no retraining, the text and LoRA+text models trained on the correct text are
evaluated with the same vector for every series: the mean CLIP text projection of the training partition. Both controls were
added after the main experiments, at the reviewers' request. We intended to run them on all
four datasets where the text-conditioned state beats the static state, but by then the PixelRec item text had been removed
from the MoTime release, so they cover Tianchi and Wiki People only.
The text and LoRA+text arms were rerun with the same settings and seeds as references;
their mean MASE is $0.006\%$ (text) and $0.005\%$ (LoRA+text) lower than in
Figure~\ref{fig:motime_methods} on Tianchi and within $0.003\%$ of it on Wiki People.

\begin{figure}[tbp]
\centering
\includegraphics[width=\linewidth]{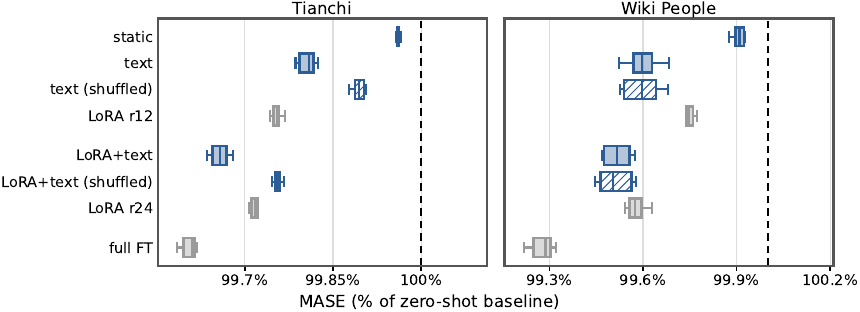}
\caption{Shuffled-text control: MASE as a percentage of the zero-shot baseline over six
seeds, lower is better. Hatched rows use the text of another series. The text and
LoRA+text rows are the reruns; all other rows are the runs of
Figure~\ref{fig:motime_methods}.}
\label{fig:shuffled_text}
\end{figure}

\begin{table}[H]
\centering
\small
\caption{Change in MASE relative to the correct text, paired by seed ($n=6$).
\emph{Shuffled}: trained and evaluated on the text of another series. \emph{Mean vector}:
the correct-text model evaluated with the same CLIP vector for every series. $\Delta$:
mean $\pm$ standard deviation of the per-seed difference; positive means higher MASE.
$p$: paired $t$-test.}
\label{tab:shuffled_text}
\begin{tabular}{lllrrcr}
\toprule
Dataset & Arm & Control & $\Delta$ MASE & $\Delta$ (\%) & seeds worse & $p$ \\
\midrule
Tianchi     & text      & shuffled    & $+0.0424 \pm 0.0038$ & $+0.088$ & 6/6 & $1.2\times10^{-6}$ \\
Tianchi     & text      & mean vector & $+0.0453 \pm 0.0038$ & $+0.094$ & 6/6 & $8.8\times10^{-7}$ \\
Tianchi     & LoRA+text & shuffled    & $+0.0472 \pm 0.0075$ & $+0.098$ & 6/6 & $2.0\times10^{-5}$ \\
Tianchi     & LoRA+text & mean vector & $+0.0687 \pm 0.0026$ & $+0.143$ & 6/6 & $1.8\times10^{-8}$ \\
\midrule
Wiki People & text      & shuffled    & $-0.0001 \pm 0.0007$ & $-0.003$ & 3/6 & $0.69$ \\
Wiki People & text      & mean vector & $-0.0003 \pm 0.0007$ & $-0.007$ & 3/6 & $0.41$ \\
Wiki People & LoRA+text & shuffled    & $-0.0003 \pm 0.0009$ & $-0.007$ & 2/6 & $0.47$ \\
Wiki People & LoRA+text & mean vector & $-0.0003 \pm 0.0007$ & $-0.007$ & 2/6 & $0.39$ \\
\bottomrule
\end{tabular}
\end{table}

On Tianchi, both controls increase MASE in every seed, with and without LoRA. The text arm
under either control remains below the static state (shuffled: $-0.0325$ MASE, Welch
$p=1.1\times10^{-5}$; mean vector: $-0.0296$, $p=3.1\times10^{-4}$), and shuffled
LoRA+text is indistinguishable from LoRA r12 ($+0.0004$, $p=0.88$). A model trained on
shuffled text does not improve when evaluated with the correct text ($-0.0002$, $p=0.85$).
On Wiki People, neither control has a measurable effect, and both keep the full gain over
the static state (shuffled: $-0.0123$, $p=3.1\times10^{-5}$; mean vector: $-0.0124$,
$p=2.7\times10^{-5}$).

\begin{table}[H]
\centering
\small
\caption{Pairwise cosine similarity between the text embeddings of distinct evaluation
series: the Qwen3 text embedding and the CLIP text projection that the
state-initialisation head receives.}
\label{tab:text_cosine}
\begin{tabular}{llrrrrr}
\toprule
Dataset & Embedding & Pairs & Min & Median & Mean & Max \\
\midrule
Tianchi     & Qwen3 ($1024$-d) & $6{,}622{,}980$ & $-0.03$ & $0.31$ & $0.32$ & $1.00^{*}$ \\
Tianchi     & CLIP ($32$-d)    & $6{,}622{,}980$ & $-0.84$ & $0.27$ & $0.26$ & $1.00^{*}$ \\
Wiki People & Qwen3 ($1024$-d) & $4{,}186$       & $-0.003$& $0.26$ & $0.26$ & $0.61$ \\
Wiki People & CLIP ($32$-d)    & $4{,}186$       & $0.53$  & $0.95$ & $0.93$ & $0.999$ \\
\bottomrule
\multicolumn{7}{l}{\footnotesize $^{*}$Some Tianchi texts occur more than once ($3{,}638$ distinct texts for $3{,}640$ series).}
\end{tabular}
\end{table}

Table~\ref{tab:text_cosine} compares the text embeddings of the two datasets. Before
alignment, the Wiki People texts are as dissimilar as the Tianchi texts (median cosine
$0.26$ vs.\ $0.31$). After the CLIP projection, the Wiki People vectors are nearly
identical (median $0.95$, minimum $0.53$), whereas the Tianchi vectors remain spread
(median $0.27$, minimum $-0.84$). To measure how much the initial state differs between
series, we split each series' state into the mean state over all evaluation series and a
series-specific deviation. For the trained text heads (mLSTM state), the summed squared
norm of the deviations is $6.1$--$6.2\%$ of the summed squared norm of the states on Wiki
People and $71$--$74\%$ on Tianchi (range over the six seeds). On Wiki People, the head
therefore writes nearly the same state into every series. The gain of the text arm over
the static state is also obtained with the same input vector for every series (mean-vector
control) and does not depend on series-specific text.

\paragraph{Interpretation.}
The two datasets give different results. On Tianchi, the
text-conditioned state uses series-specific information: removing it raises MASE by
$0.09$--$0.14\%$ in every seed, and without it LoRA+text no longer improves on a LoRA
adapter of the same rank. Part of the Tianchi gain is nevertheless not tied to the series'
own text: $40$--$43\%$ of the text arm's improvement over the static state remains under
either control. On Wiki People, the advantage of the text arm over the static state does
not come from exogenous information. The texts are distinct, but the CLIP projection maps
them to nearly the same vector, so the series-specific information is lost before it
reaches the state, and the head writes almost the same state into every series. The text
head then acts as a shared static state. It differs from the learnable static state in
being shared across layers, its initialisation and its parameterisation (a
LayerNorm-normalised projection); which of these accounts for its lower MASE was not tested. The Wiki People result in
Figure~\ref{fig:motime_methods} is therefore not evidence that the state carries
exogenous text.

\end{document}